\documentclass[10pt, a4paper]{article}
\usepackage{IOA}
\usepackage{blindtext}
\usepackage{layout}

\usepackage{graphicx}
\usepackage{amsmath}								
\usepackage{amssymb}								
\usepackage{float}
\usepackage{tikz}
\usepackage{array}
\usepackage{makecell}
\usepackage[space]{grffile}
\usepackage{tikz}
\usetikzlibrary{positioning,calc,matrix,arrows}
\usepackage[backend=biber, sorting=none]{biblatex}
\title{Additional Representations for Improving Synthetic Aperture Sonar Classification Using Convolutional Neural Networks}

\firstauthor{ID Gerg}
\secondauthor{DP Williams}
\thirdauthor{}
\fourthauthor{}
\fifthauthor{}
\sixthauthor{}

\firstaffl{Pennsylvania State University Applied Research Laboratory (PSU-ARL), State College, Pennsylvania, USA}
\secondaffl{Centre for Maritime Research \& Experimentation (CMRE), La Spezia, Italy}
\thirdaffl{}
\fourthaffl{}
\fifthaffl{}
\sixthaffl{}

\begin{document}

	\maketitle 
	\makeauthors

\section{Introduction}
Object classification in synthetic aperture sonar (SAS) imagery is usually a data starved and class imbalanced problem. There are few objects of interest present among much benign seafloor. Despite these problems, current classification techniques discard a large portion of the collected SAS information.  In particular, a beamformed SAS image, which we call a single-look complex (SLC) image, contains complex pixels composed of real and imagery parts.  For human consumption, the SLC is converted to a magnitude-phase representation and the phase information is discarded.  Even more problematic, the magnitude information usually exhibits a large dynamic range ($>$80dB) and must be dynamic range compressed for human display.  Often it is this dynamic range compressed representation, originally designed for human consumption, which is fed into a classifier.  Consequently, the classification process is completely void of the phase information.

\begin{figure}[h]
	\begin{center}
		\newcolumntype{C}{>{\centering\arraybackslash} m{2cm} }
		\newcolumntype{D}{>{\centering\arraybackslash} m{3cm} }
		\begin{tabular}{C D D D}
			& Target & Clutter 1 & Clutter 2 \\
			Dynamic Range Compressed Magnitude & 
			\includegraphics[scale=0.25]{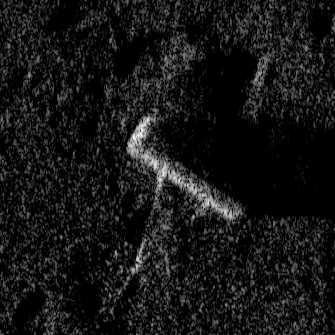} &
			\includegraphics[scale=0.25]{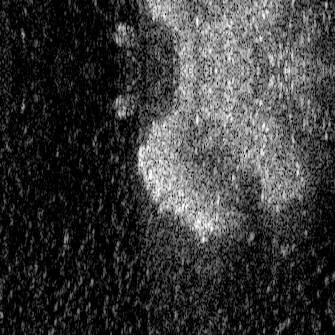} &
			\includegraphics[scale=0.25]{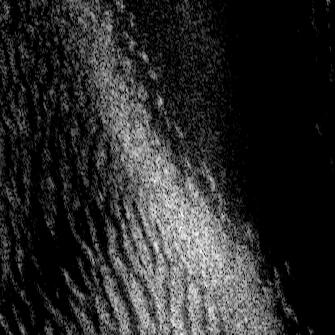}	 \\		
			
			Phase &
			\includegraphics[scale=0.25]{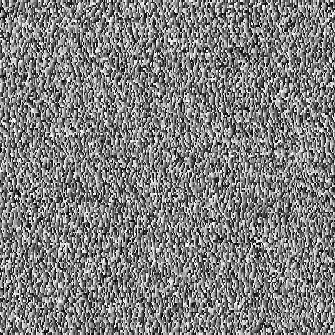} &
			\includegraphics[scale=0.25]{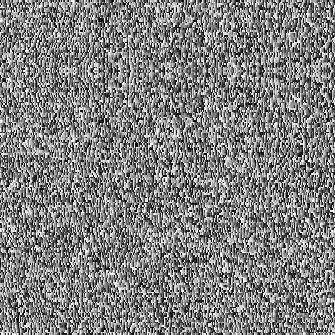} &
			\includegraphics[scale=0.25]{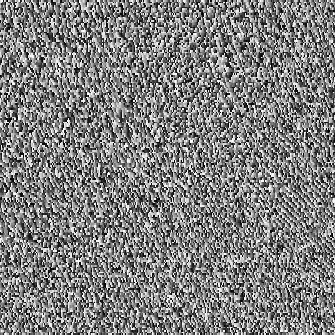}	 \\	
			
			Unwrapped Phase &
			\includegraphics[scale=0.25]{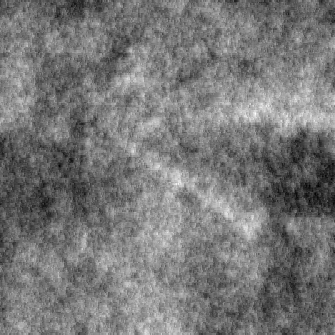} &
			\includegraphics[scale=0.25]{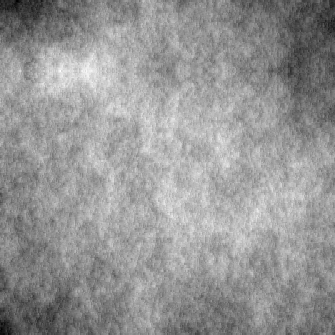} &
			\includegraphics[scale=0.25]{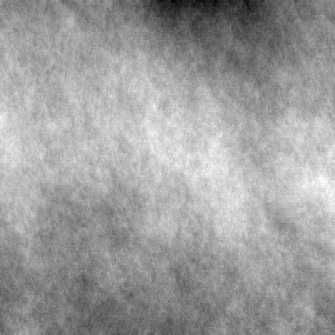}	 \\		
		\end{tabular}
	\end{center}
	\caption{Sample images from our dataset. Top row is dynamic range compressed (DR) magnitude imagery, middle row is raw phase, and bottom row is 2D unwrapped phase and de-trended \cite{ghiglia1994robust}. Left-most image is a target and remaining images are clutter. }
	\label{fig:dataset_sample_images}
\end{figure}

The discarded phase information from an SLC was recently shown to have utility in SAS classification \cite{williamsexploiting}. Figure 	\ref{fig:dataset_sample_images} shows an example of how the phase information can be processed to reveal features correlated with the magnitude image. This work leads us naturally to ask two questions: (1) \emph{What representations can be derived from the SLC, specifically the phase representation, to improve ATR performance?} and (2) \emph{Are there additional representations outside of the SAS phenomenology which could improve classification?}  In this work, we will answer both questions.

Specifically, this work provides three contributions:
\begin{enumerate}
	\item We will show that convolutional neural networks (CNNs) discover features from SLC imagery which are human interpretable.
	\item We will demonstrate that augmenting classifier input with the power spectral density (PSD) of the SLC improves classification performance over input of magnitude imagery alone.
	\item We will show that a pre-trained off-the-shelf (OTS) CNN trained on photographs can be fine-tuned to classify SAS images with good performance.
\end{enumerate}

The remainder of the paper is organized as follows: Section 2 will describe the experimental setup and data, Section 3 will present the classification results, Section 4 will provide further analysis of the results by examining the latent space of the trained classifiers, and finally, in Section 5, we will present our conclusions.

\section{Experiment Setup} 
In this section, we will discuss our experimental approach, justifications for our setup, and details about the dataset used.

\subsection{Problem \& Approach} 

We seek to determine if representations derived from a SAS SLC image can improve neural network classification performance. Likely, the same set of features cannot be used for all representations because of the different physics captured.  Furthermore, the optimal set of features for these different representations is unclear. To mitigate this \emph{feature engineering} step, we choose a convolutional neural network (CNN) because of its ability to generate features automatically. 

We accomplish the task by creating a fundamental CNN architecture derived from \cite{williamsdemystifying} and compare performance among combinations of input representations derived from the SLC. We train each classifier using the same data and evaluate the results using the receiver-operating-characteristic (ROC) area-under-the-curve (AUC) metric because of its indifference to class imbalance.  We then use significance testing to evaluate the results.

\subsection{Data}
The data we used in the experiment was collected from the CMRE MUSCLE SAS system.  MUSCLE is a thirty-two channel side looking SAS mounted on a Bluefin-21 unmanned underwater vehicle. The SAS operates at center frequency 300kHz with a bandwidth of 60kHz.  The nominal imaging range of the system  is 140 meters and maximum water depth is eighty meters. The system operates at a fixed ping rate and travels at a nominal velocity of 3.5 knots.

The dataset contains imagery from thirteen naval exercises, we call trials, over a variety of environments. The imagery was parsed into smaller image chips using the Mondrian detector \cite{williams2018mondrian} as a pre-screener. The total number of chips available was approximately 56,000 and each chip measures 5m x 5m. We split the chips into training and test sets. The split was made chronologically by assigning chips collected during the first half of the trials to the training set and the remainder to the test set.  This resulted in a relatively equal split of training and test samples. Table \ref{tab:dataset_breakdown} shows details of the train/test split and Table \ref{tab:dataset_breakdown_by_trial} shows proportion by trial for the test set.

\begin{table}
	\centering
	\caption{Breakdown of the image chips used to train the classifier. C/T is clutter/target ratio.}
	\begin{tabular}[h]{l c c c c}
		& No. Clutter & No. Targets	 & C/T & Exercise Dates (No. Experiments)\\
		\hline
		Training	& 29,280             &  2,912          & $\approx$ 10 & 2008-2013 (8) \\
		Test		& 23,099             &  1,627          & $\approx$ 14 & 2014-2017 (5)\\
	\end{tabular}
	\label{tab:dataset_breakdown}
\end{table}

\begin{table}
	\centering
	\caption{Breakdown of the image chips used in the test set by trial.}
	\begin{tabular}[h]{l c}
		Trial Name & Proportion of Test Samples \\
		\hline
		ONM1	& 0.93 \% \\
		GAM1	& 1.2 \% \\
		TJM1	& 9.8 \% \\
		NSM1	& 28 \% \\
		MAN2    & 60 \% \\
	\end{tabular}
	\label{tab:dataset_breakdown_by_trial}
\end{table}

We examine classifier performance using combinations of three representations derived from the SLC: dynamic range compressed magnitude, phase, and 2D power spectral density (PSD).  The magnitude image was dynamic range compressed from the SLC using a proprietary algorithm.  The phase was computed pointwise from each complex pixel and mapped to the range $[0, 2\pi).$ Finally, the 2D PSD is computed as the power of the DC-centered 2D Fourier transform of the the SLC.

Finally, we utilized a pre-trained off-the-shelf (OTS) convolutional neural network, a VGGnet \cite{simonyan2014very}, for the task of evaluating how well photographic features can improve classification performance.  The network was trained on the ImageNet dataset which consists of 1.1 million photographs categorized into 1000 classes.

\subsection{Convolutional Neural Network Architecture}

\begin{figure}
	\centering
	\newcolumntype{C}{>{\centering\arraybackslash} m{7cm} }
	\newcolumntype{C}{>{\centering\arraybackslash} m{7cm} }
	\begin{tabular}{C C}
		
		\begin{tikzpicture}
		[
		graynode/.style={, draw=black!90, fill=black!20, very thick, minimum size=5mm},
		graynodecircle/.style={circle, draw=black!90, fill=black!20, very thick, minimum size=5mm},
		whitenode/.style={rectangle, draw=black!90, very thick, minimum size=5mm},
		inout/.style={rectangle, draw=black!0, very thick, minimum size=5mm},
		scale=0.6, every node/.style={scale=0.6}
		]
		
		\begin{scope}[node distance=3mm and 10mm] 
		\node[graynode] (conv1) {Conv, 8, 8};
		\node[graynode] (conv2) [above=of conv1] {Avg Pool, 4};
		\node[graynode] (conv3l) [above=of conv2, xshift=-1.25cm] {Conv, 10, 6};
		\node[graynode] (conv3r) [right=of conv3l, xshift=-1cm] {Conv, 10, 1};
		\node[graynodecircle] (conv4) [above=of conv3l,xshift=+1.25cm] {+};
		\node[graynode] (avgpool2) [above=of conv4] {Avg Pool, 4};
		
		\node[graynode] (conv12l) [above=of avgpool2, xshift=-1.25cm] {Conv, 12, 6};
		\node[graynode] (conv12r) [right=of conv12l, xshift=-1cm] {Conv, 12, 1};
		
		\node[graynodecircle] (add2) [above=of conv12l, xshift=+1.25cm] {+};
		\node[graynode] (flatten) [above=of add2] {Flatten};
		\node[whitenode] (dropout) [above=of flatten] {Dropout};
		\node[whitenode] (fc) [above=of dropout] {Fully Connected};
		
		\node[inout] (output) [above=of fc] {Output};
		\node[inout] (input) [below=of conv1] {Input};
		\end{scope}
		
		\draw[->] (input) -- (conv1);
		\draw[->] (conv1) -- (conv2);
		
		\draw[->] (conv2) to (conv3l);
		\draw[->] (conv2) -- (conv3r);
		
		\draw[->] (conv3l) to (conv4);
		\draw[->] (conv3r) -- (conv4);
		\draw[->] (conv4) -- (avgpool2.south);
		\draw[->] (avgpool2) -- (conv12l);
		\draw[->] (avgpool2) -- (conv12r);
		\draw[->] (conv12r) -- (add2);
		\draw[->] (conv12l) -- (add2);
		\draw[->] (add2) -- (flatten);
		\draw[->] (flatten) -- (dropout);
		\draw[->] (dropout) -- (fc);
		\draw[->] (fc) -- (output);
		
		\end{tikzpicture} &
		
		\begin{tikzpicture}
		[
		graynode/.style={, draw=black!90, fill=black!20, very thick, minimum size=5mm},
		graynodecircle/.style={circle, draw=black!90, fill=black!20, very thick, minimum size=5mm},
		whitenode/.style={rectangle, draw=black!90, very thick, minimum size=5mm},
		inout/.style={rectangle, draw=black!0, very thick, minimum size=5mm},
		]
		
		\begin{scope}[node distance=3mm and 3mm ] 
		
		\node[inout](altinput)[align=center]{Alternate Representation \\ Input};
		\node[inout](maginput)[left=of altinput, align=center]{Magnitude Input\\};
		
		\node[graynode](maginputblock)[above=of maginput, align=center]{Magnitude Net\\};
		\node[graynode](altinputblock)[above=of altinput, align=center]{Alternate Representation \\Net};
		
		\node[whitenode](concat) [above=of altinputblock, xshift=-2.25cm] {Concatenate};
		\node[whitenode](dropout) [above=of concat] {Dropout};
		\node[whitenode](fc) [above=of dropout] {Fully Connected};
		
		\node[inout] (output)[above=of fc] {Output};
		\end{scope}
		
		\draw[->] (maginput) -- (maginputblock);
		\draw[->] (altinput) -- (altinputblock);
		
		\draw[->] (maginputblock) to (concat);
		\draw[->] (altinputblock) -- (concat);
		\draw[->] (concat) -- (dropout);
		\draw[->] (dropout) -- (fc);
		\draw[->] (fc) -- (output);
		
		\end{tikzpicture} 
	\end{tabular}
	
	\caption{Network architecture used as the building block for creating the nets. The gray shaded path in the network on the left is replicated for each input representation.  The diagram as shown is verbatim the architecture for the Magnitude-only input configuration. Additional inputs are added as parallel paths as shown on the right. The convolutional layers are denoted \{Conv, \emph{number of filters}, \emph{kernel size}\} and average pooling layers denoted as \{Avg Pool, \emph{pool size\}}}
	\label{fig:network_arch}
\end{figure}
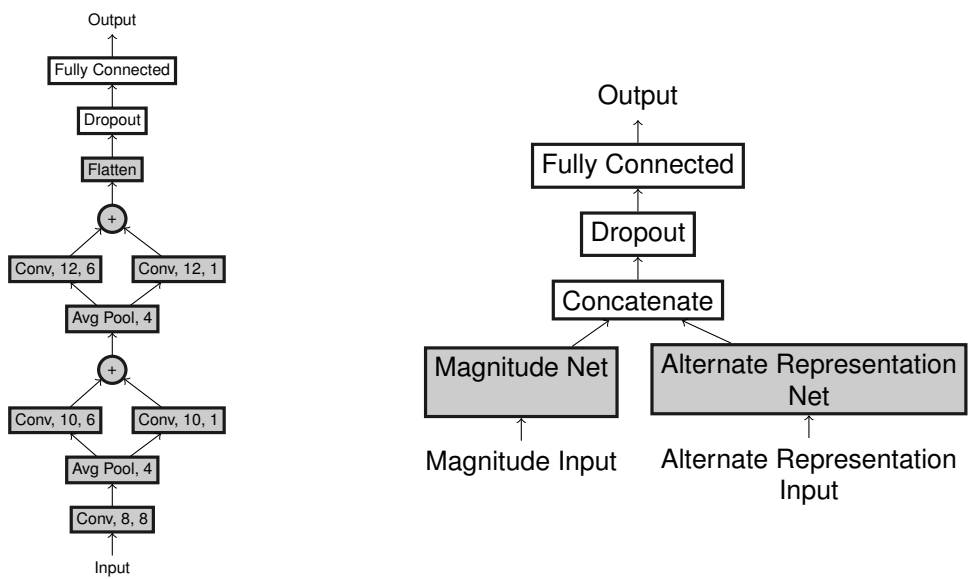

For the representations derived from the phase of the SLC, we use a fundamental network architecture and duplicate it as parallel paths concatenating their outputs into a dropout and fully connected layer. This architecture is shown in Figure \ref{fig:network_arch} and is an improvement on \cite{williamsdemystifying}; we add skip layers to improve convergence \cite{li2017visualizing} and use the ReLU activation everywhere. Each net contains approximately 11k free parameters. For all networks, the inputs are scaled to the range [-1, 1].

The concatenation among parallel paths was done by flattening the final convolution layer output and concatenating along the dominant axis. The concatenation was fed into a dropout layer and then into a fully connected layer with a single sigmoid output.

The pre-trained network was a VGGnet trained on Imagnet. We only evaluated magnitude chips using this setup. VGGnet expects a three-channel color input but the chips are single-channel grayscale.  We mitigated this issue by duplicating the grayscale values for each pixel in the input. Only the convolutional layers were borrowed from this architecture as we added flattening, dropout, and fully connected layers consistent with the architecture of Figure \ref{fig:network_arch}.

Overall, we evaluated seven input configurations: (1) Magnitude-only serving as our reference for statistical testing, (2) Phase-only, (3) PSD-only, (4) Magnitude \& Phase, (5) Magnitude \& PSD, (6) Magnitude \& Phase \& PSD, and (7) Magnitude-only but using the OTS pre-trained network.

\subsection{Training}

We use binary cross-entropy as the training loss and RMSProp \cite{ruder2016overview} as the optimizer for all experiments. We conduct the experiments on a single NVIDIA GTX-960 graphics processing unit (GPU). 

Hyper-parameters are shown to have a significant impact on classifier performance \cite{henderson2017deep}; changing random seeds can results in significant differences all other things held constant. So, we desire to have as few hyper-parameters as possible in order to fairly evaluate the networks.  However, there are a few hyper-parameters which we cannot avoid: the learning and dropout rates. 

The learning rate modulates the gradient magnitude in searching for a solution to the network.  Setting this value too low results in slow convergence and setting it too high results in chaotic behavior. For our custom networks, we set this parameter by conducting a pilot study on a subset of the data.  We first set the learning rate to $1e-6$ and then train. Next, we increase the learning rate by a factor of ten and repeat the process.  We select the largest learning rate which gives monotonic convergence against the training set.  The resulting learning rate was the same for all networks: $1e-3$.

We train the OTS network using a fine-tuning approach.  For the first epoch, we freeze all the convolutional weights and execute the learning procedure.  For all further epochs we unfreeze all the weights. We find this methodology prevents large gradients, due to random initialization of the fully connected layer, from influencing the adaption in a negative fashion. We use a learning rate of $1e-6$ for the initial epoch and $1e-5$ for all others.

SAS datasets are typically class imbalanced.  The ratio of background-to-target is  generally on the order of $1000:1$.  We greatly reduced this ratio by using a detector as a pre-screener.  This improved the class imbalance to $10:1$. Despite the 100x reduction by the detector though, our dataset is still quite imbalanced.  

We mitigate the dataset imbalance using the approach of \cite{wallace2011class}. In this work, the authors show training a classifier by sub-sampling with replacement yields a biased result.  The authors propose to even-sample the classes which removes the bias, but increases the variance.  The variance is removed by averaging several models.  We mimic this philosophy in our CNN by even-sampling the classes and using dropout \cite{srivastava2014dropout}.  Dropout works by randomly setting a portion of the model weights to zero thereby forming random submodels each mini-batch where a mini-batch is the subset of training data used to compute the model error and update the weights. During test time, dropout is disabled and the result is an averaging of the submodels. 

We even-sample our classes in two ways.  First, we augment the minority class (i.e. target class) by adding vertically flipped versions of each chip mimicking the sonar traveling in the opposite direction.  And second, we augment the minority class by selecting random 4m x 4m crops of the original 5m x 5m chips.

CNNs are time-consuming to train from scratch.  We cannot simply train up several dozen models from scratch and average their results to mitigate the large variance resulting from our even-sampling.  However, we approximate an ensemble of networks by using dropout. Dropout has several additional benefits including preventing co-adaptation between nodes and mitigating overfitting resulting from the large difference in fan-in \& fan-out between the concatenation and fully connected layers. We determine the proportion of dropout to use by doing small pilot studies on a subset of the data and selecting the best results from the set of proportions $\{0\%, 50\%, 66\%, 75\%, 90\%\}$. 

Additionally, we prevent overfitting by using early stopping during training. We stop the training procedure when the maximum test set AUC did not occur in the last twenty epochs.

\subsection{Statistical Analysis}

We demonstrated statistical significance by using the Wilcoxon signed-rank (WSR) test and boostrapping.  We use a non-parametric test since our metric distribution is unknown and not Gaussian distributed. Recall, AUC is our performance metric and is bounded to [0,1].  The distribution of a bounded random variable cannot be Gaussian by definition.  However, a Gaussian distribution can closely approximate a bounded random variable especially when not near the bounds.  However, in this case, we observe our AUCs are close to one.  Therefore, a Gaussian assumption is likely to be a poor choice. We complete the WSR test by forming an ensemble of AUCs by performing bootstrapping one-hundred times.  

The null hypothesis for our experiment is there is no difference in AUC between the Magnitude-only configuration and every other. We set the level of significance to be $p<1e-3$. This is a six-way comparison so we apply a Bonferroni correction to the p-value to account for accidental significance from multiple comparisons.

\section{Results}

\begin{figure}[h]	
	\centering
	\includegraphics[scale=0.75]{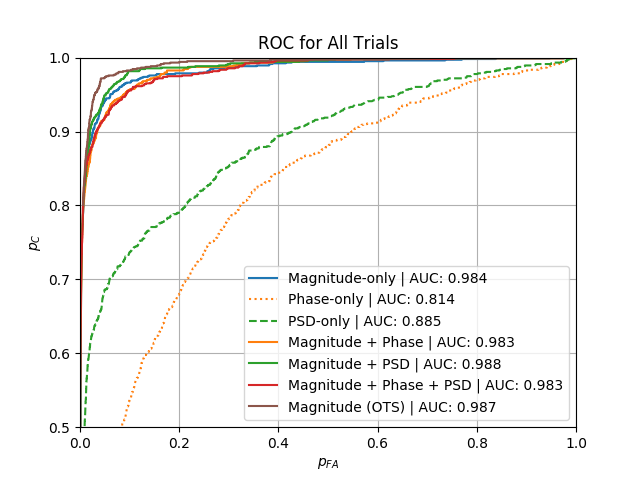} 
	\caption{ROCs for each of the representations computed on all the data.}
	\label{fig:overall_rocs_for_each_representation}
\end{figure}

\begin{figure}
	\centering
	\includegraphics[scale=0.75]{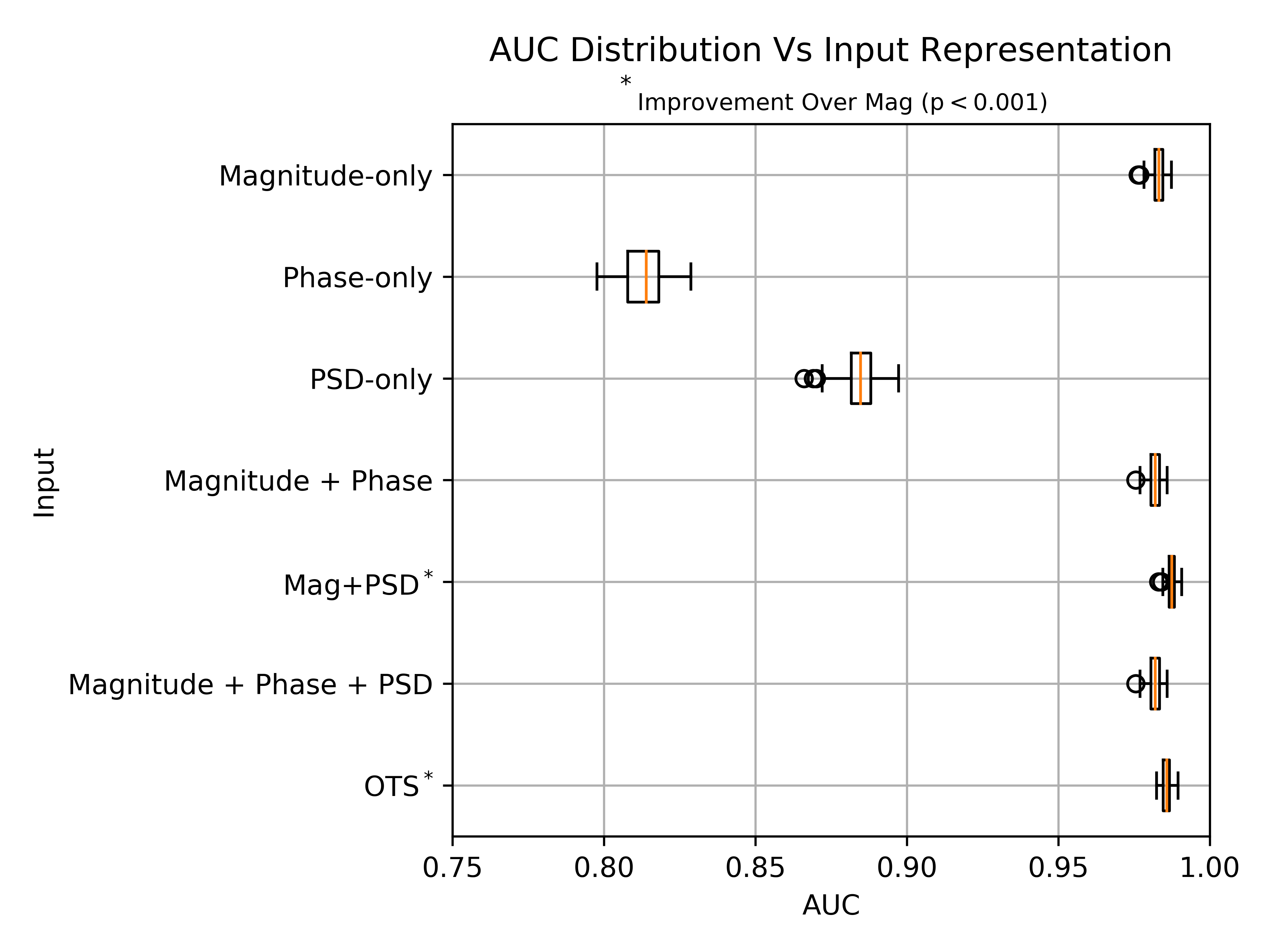} 
	\caption{Box and whisker plot of the AUCs for each input type.  Improved AUC with respect to Magnitude-only input denoted with asterisk (*). Each box extends from the lower to upper quartile and the vertical orange line represents the median. }
	\label{fig:auc_box_and_whisker}
\end{figure}

The ROCs for each configuration are shown in Figure \ref{fig:overall_rocs_for_each_representation}. We saw from this plot that two configurations generally perform better than Magnitude-only, Magnitude + PSD and Magnitude OTS.  We found these differences to be statistically significant. Figure \ref{fig:auc_box_and_whisker} shows the distributions of AUC from the bootstrapping procedure and notes the configurations with improved AUC compared to the Magnitude-only configuration. It is worth noting that we get a similar, non-trivial AUC for the Phase-only configuration which is consistent with \cite{williamsexploiting}.

We saw varying performance of the configuration across environments; however, configurations with Magnitude combined with PSD or OTS pre-trained weights do best. Results shown in Table \ref{table:auc_by_trial}.

\begin{table}	
	\caption{AUCs for each classifier by trial.}
	\centering
	\resizebox{\textwidth}{!}{%
		\begin{tabular}{l c c c c c c c c c }
			Trial & Proportion & Magnitude & Phase & PSD & Mag+Phase & Mag+PSD & Mag+Phase+PSD & Mag (OTS) \\	
			\hline
			ONM1 	& 0.93\%	& 0.992 & 0.749 & 0.611 & 0.976 & 0.976 & \textbf{0.993}  & 0.985 \\
			GAM1 	& 1.2\%		& 0.988 & 0.726 & 0.969 & 0.992 & 0.967 & 0.990 & \textbf{1.000} \\
			TJM1 	& 9.8\%	& 0.995 & 0.779 & 0.971 & 0.996 & 0.996 & 0.995 & \textbf{0.997} \\
			NSM1 	& 28\%	& 0.946 & 0.647 & 0.903 & 0.954 & \textbf{0.981} & 0.955  & 0.981 \\
			MAN2 	& 60\%	& 0.981 & 0.897 & 0.923 & 0.982 & 0.983 & 0.980 & \textbf{0.986} \\
			\hline
			All 	& 100\%   & 0.984 & 0.814 & 0.885 & 0.984 & 0.988 & 0.983  & \textbf{0.989} \\
	\end{tabular}}
	\label{table:auc_by_trial}
\end{table}

\section{Discussion}

The results of Section 3 demonstrate improved AUC when training SAS imagery with either additional information (as in the VGGnet pre-trained from photographs) or usually discarded information (e.g. improvement by adding PSD to the input). Up to this point, we have treated the networks as universal approximators, black-boxes if you will, with little regard of the internal mechanics at work.  In this section, we examine the network's learned weights and latent space to understand how we might improve upon our current results.

\subsection{Network Weights}
\begin{figure}[h]
	\centering
	\begin{tabular}{c}
		\includegraphics[trim={1.1cm 3cm 9.5cm 3.0cm},clip,scale=0.75]{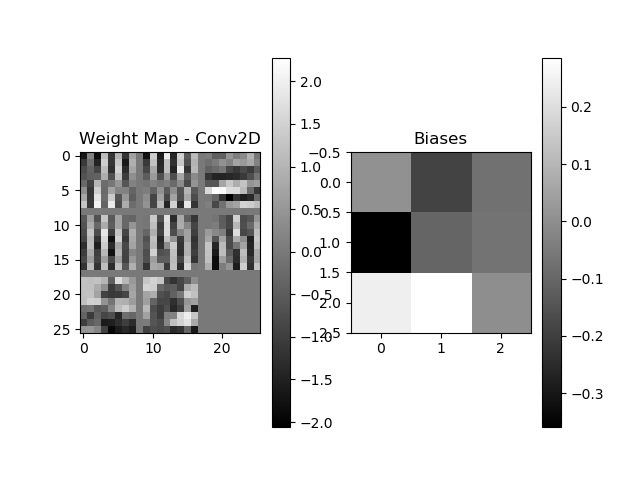} \\
	\end{tabular}	
	\caption{The convolutional filters of the first convolutional block for the Phase-only representation.}
	\label{fig:cnn_layers_phase}
\end{figure}

We examined the network weights of Phase-only representation and noticed a striking reversal pattern occurring in the first convolutional layer.  We then examined the phase data and noticed it contained phase wrapping artifacts in similar patterns as these weights. We hypothesize some type of phase unwrapping is occurring at the lowest layer; see Figure \ref{fig:cnn_layers_phase} for a plot of the convolutional weights.

\begin{figure}
	\centering
	\begin{tabular}{c c c }
		\includegraphics[trim={50px 70px 20px 70px},clip,scale=0.33]{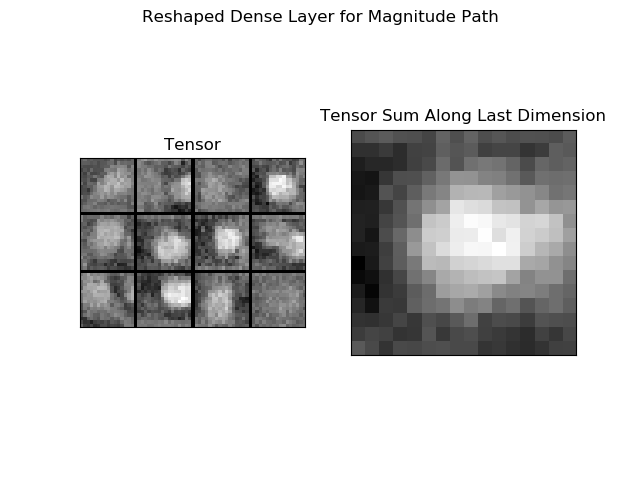} & 				\includegraphics[trim={50px 70px 20px 70px},clip,scale=0.33]{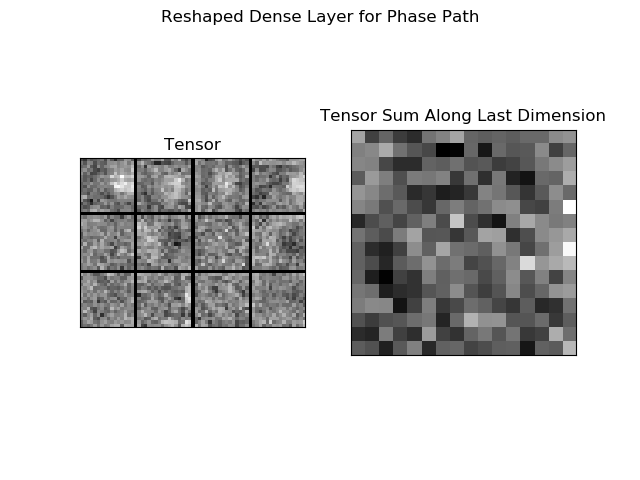} &
		\includegraphics[trim={50px 70px 20px 70px},clip,scale=0.33]{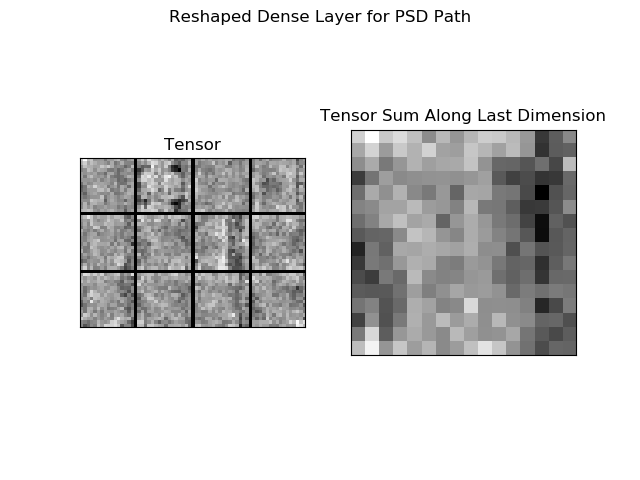} \\
		(a) Magnitude & (b) Phase & (c) PSD\\		
	\end{tabular}
	\caption{Un-flattened weights of the fully connected layer arranged according to their corresponding spatial location of the incoming convolutional output.}
	\label{fig:dense_layer_weights_spatially}
\end{figure}

Similarly, we analyzed the weights of the fully connected layer. These weights correspond to the flattened tensor of the last convolutional layer output.  Unlike the weights examined in the case of phase above, these weights correspond to spatial locations of network output.  Another way to say it, the convolutional weights are translation invariant whereas the fully connected weights are spatially dependent.  We un-flatten the weights and arrange them according to their convolutional output as well as coherently sum them as a function of spatial location. Figure \ref{fig:dense_layer_weights_spatially} shows these maps.  We see in the magnitude path that the weights mimic highlight-shadow arrangements which is typically observed in SAS imagery.  We see similar patterns in the phase but not nearly as strong.  In the PSD representation, we see areas of high weighting in the leftmost corners of the k-space.  In examining the PSD input, we see strong areas of texture in these regions.  Further work will investigate this phenomena.

\subsection{Mutual Information of the Feature Spaces}
Part of the training procedure of CNNs is creating the feature space. We wondered how the feature space differs between networks trained with alternate representations alone (i.e. PSD-only or Phase-only) versus those trained with augmentation of the magnitude image (i.e. Magnitude+PSD and Magnitude+Phase).  Specifically, we wondered how the feature discovery process was modulated when combined representation are used as input to the networks. To measure this, we examined the mutual information (MI) of features output by the last convolutional layer between the exclusive inputs and augmented inputs. The mutual information was measured using Krakov's method \cite{kraskov2004estimating} using the recommended parameter of $k=3$. Mutual information estimates become biased in high dimensions so we projected the feature space down to ten dimensions using principal component analysis (PCA) before measuring MI.

To measure the MI between single-input nets, we performed the procedure above using the output of the last convolutional layer from a subset of the input. For example, to measure the feature space MI between the Magnitude-only network and the Phase-only network, we input the same subset of the data to each network, compute PCA on the feature vectors output by the last convolutional layer, and then measure the MI between these point clouds.  For single input networks, we found the most amount of MI existed between the Magnitude and PSD feature spaces.

We also measured the MI between feature spaces of the different inputs when trained together.  To perform this measurement, we use the same procedure as the single input nets but now compare the points from the last convolutional layer of each parallel path. Overall, we see an increase in MI when multiple inputs are trained together than apart. This suggests the feature spaces are more statistically dependent when trained together than apart. Additionally, we see the largest increase in MI when the Magnitude and PSD representations are trained together.  This fact, and increased AUC, suggests a type of synergy occurring between the representations when trained together than apart.

\begin{figure}	
	\centering
	\includegraphics[scale=0.55]{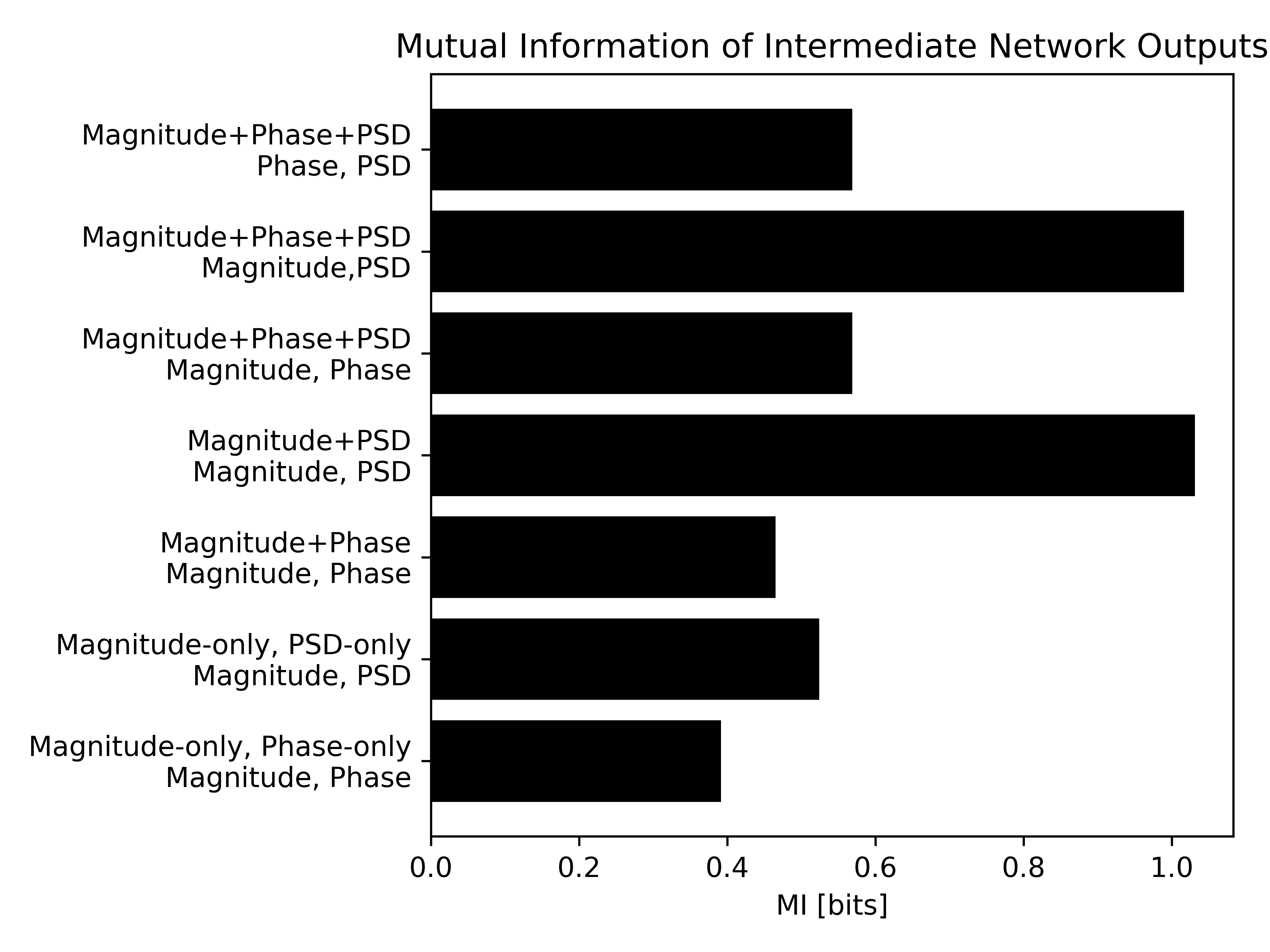} 
	\caption{Mutual information (MI) of internal representations. The top label of each bar denotes the network source(s) and the bottom label denotes the internal representations used to measure the MI.}
	\label{fig:mis}
\end{figure}

\subsection{Unsupervised Clustering by Trial in the Feature Space}
\begin{figure}[h]
	\centering
	\includegraphics[scale=0.95]{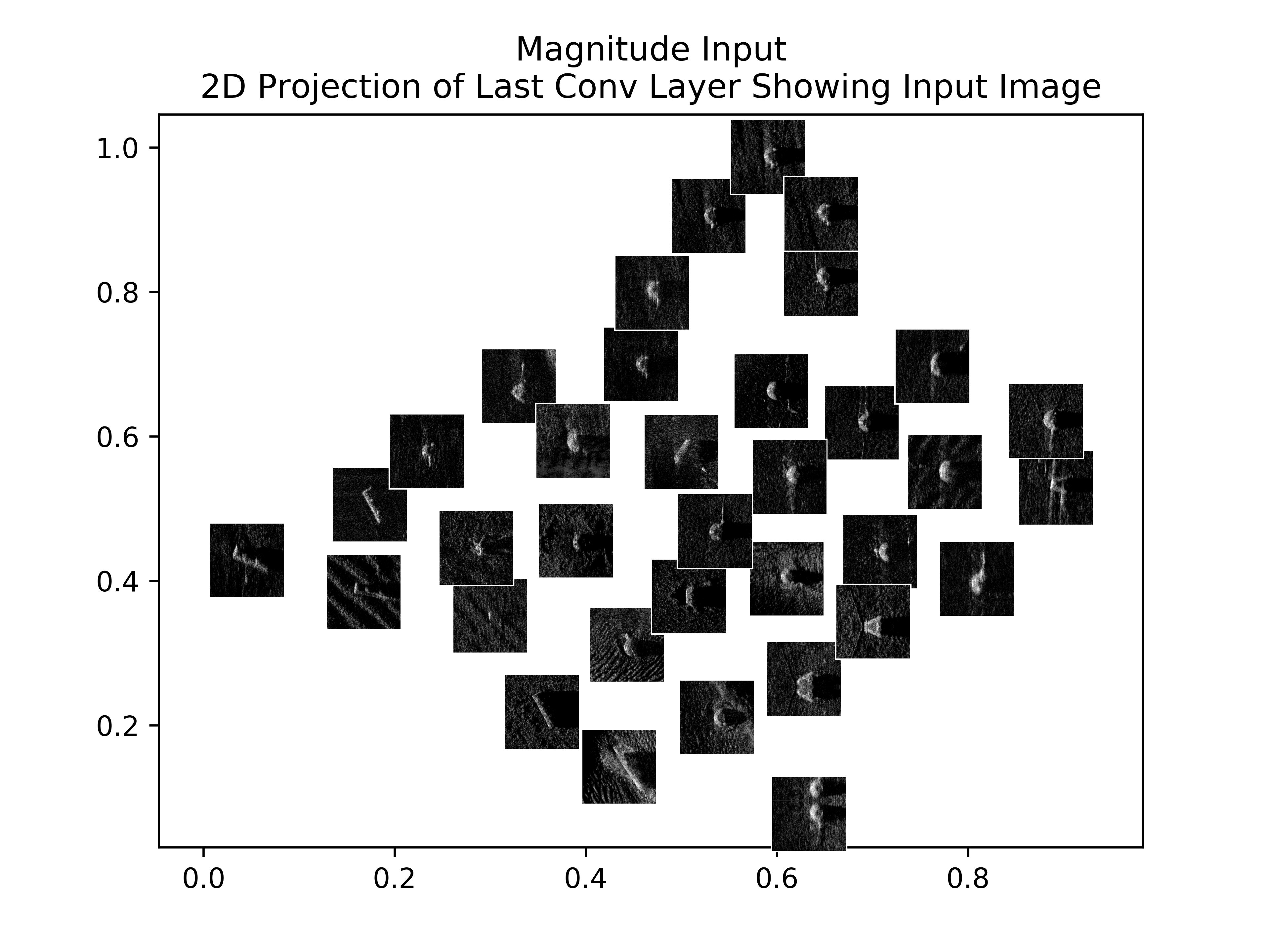}
	\caption{Magnitude images arranged in t-SNE projected feature space for Magnitude-only network.}
	\label{fig:tsne_magonly}
\end{figure}

We examined the 2D projection of the feature spaces from the output of the last convolutional layer.  We projected the data points to $\mathbb{R}^{2}$ using t-SNE \cite{maaten2008visualizing} and plotted their corresponding representation.  An example output of this procedure for Magnitude-only input is shown in Figure \ref{fig:tsne_magonly}. The plot shows feature organization by target/object type with spherical shapes near the top, wedge shapes near the bottom right, and cylindrical shapes on the bottom left. 

\begin{figure}[h]
	\centering
	\includegraphics[scale=0.75]{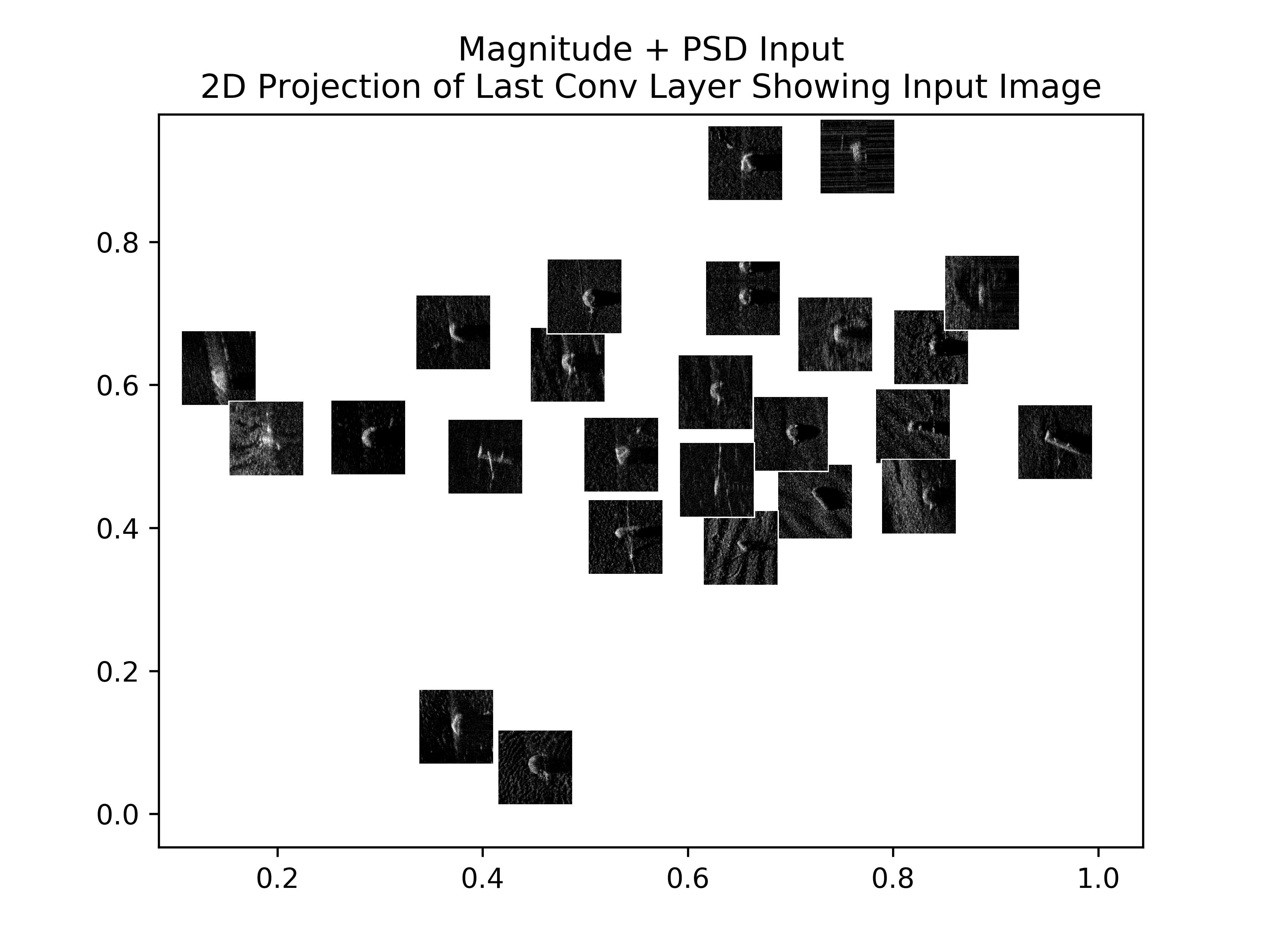}
	\includegraphics[scale=0.75]{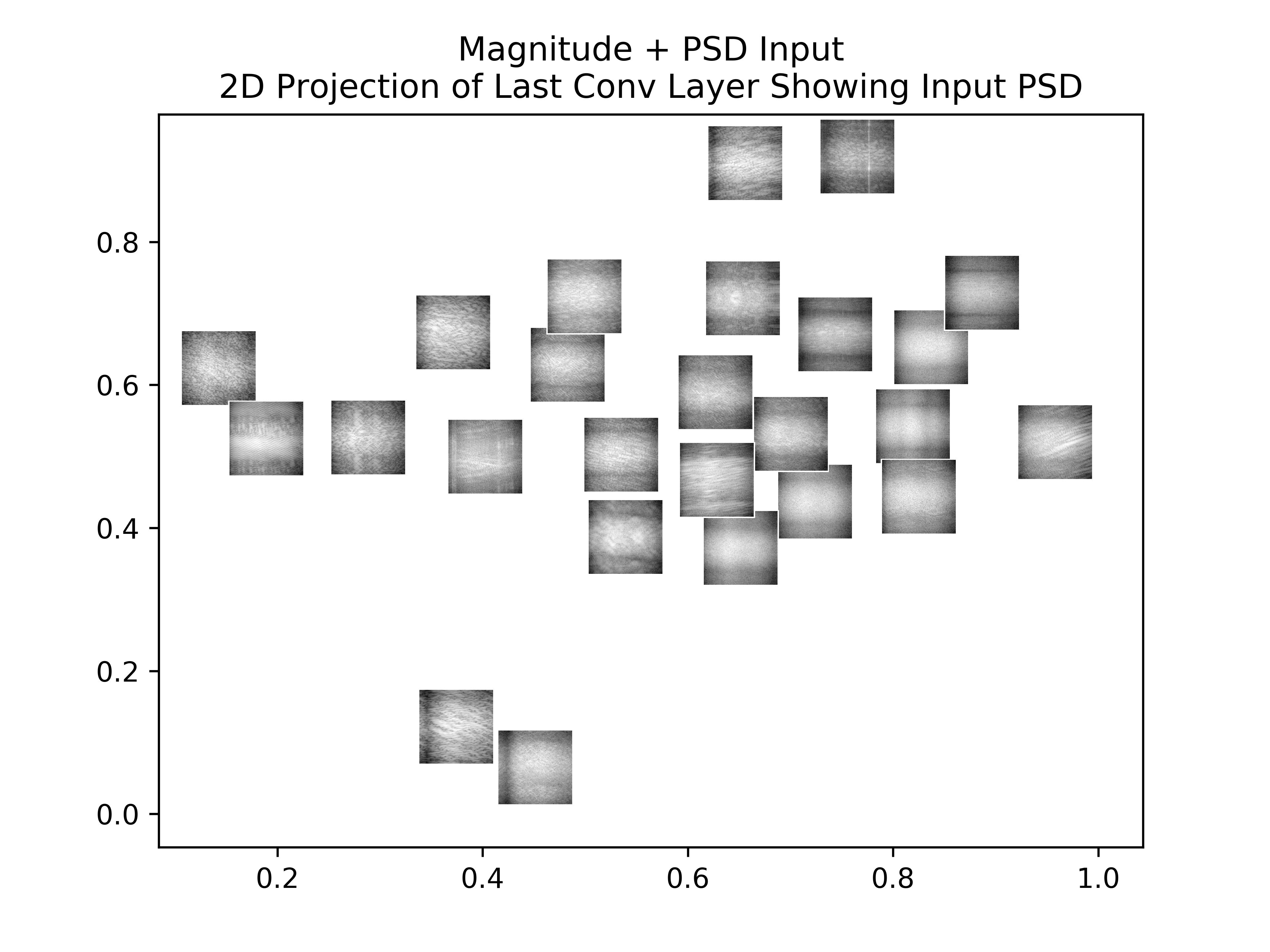}	
	\caption{Magnitude (top) and PSD (bottom) images arranged in t-SNE projected feature space for Magnitude+PSD network.}
	\label{fig:tsne_mag_mag_psd}
\end{figure}

\begin{figure}[h]
	\centering
	\includegraphics[scale=0.55]{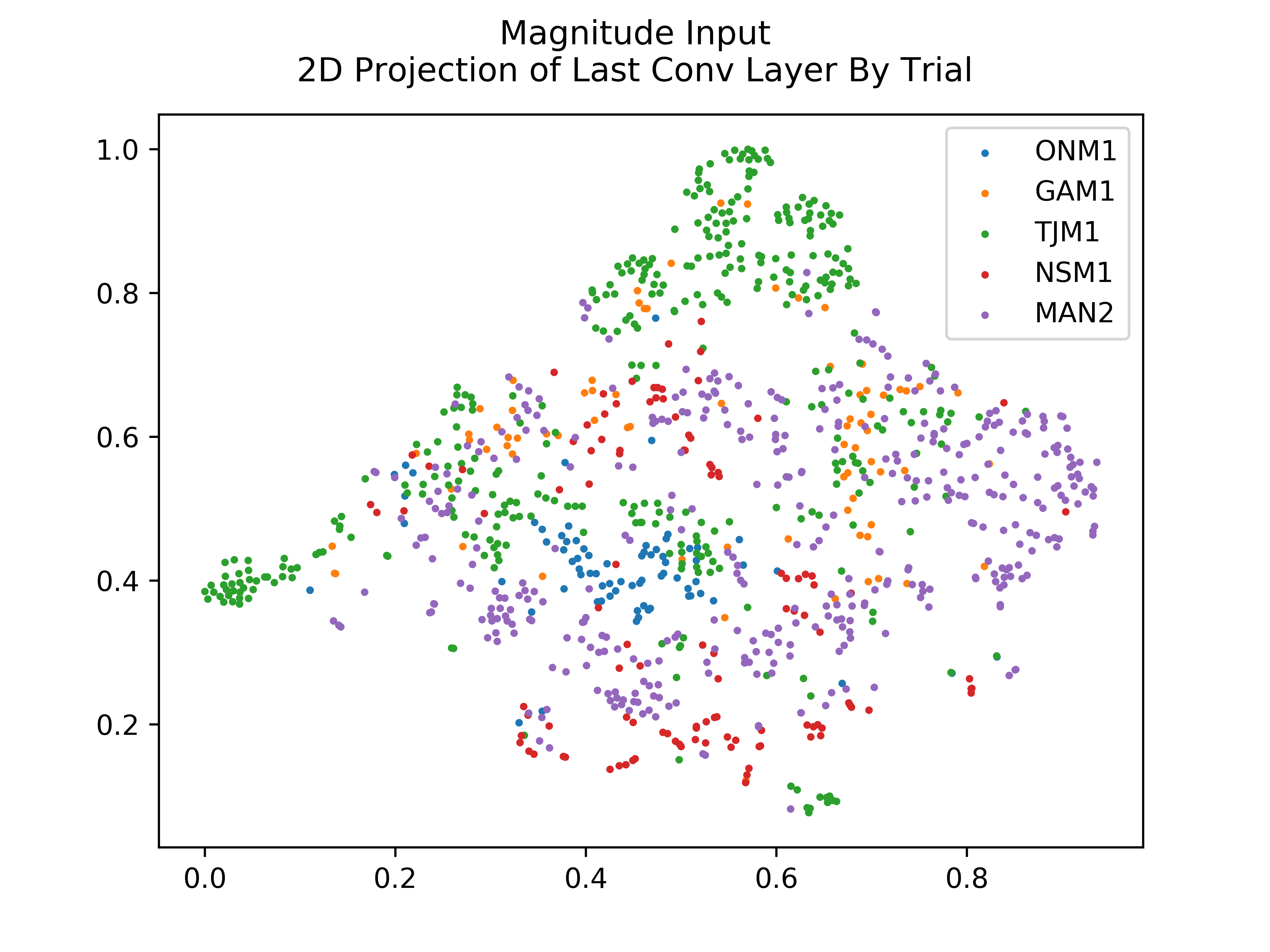}
	\includegraphics[scale=0.55]{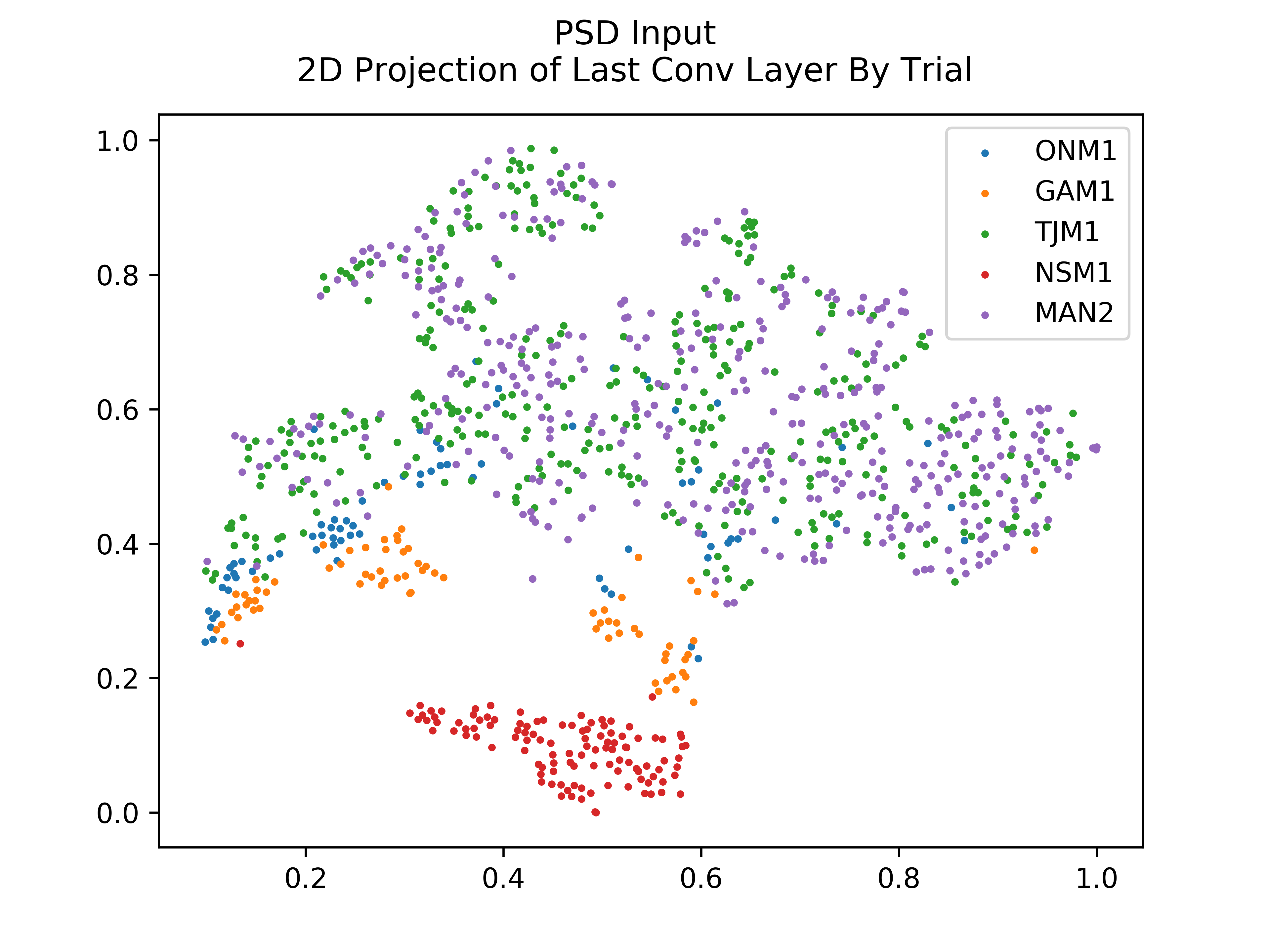}	
	\includegraphics[scale=0.55]{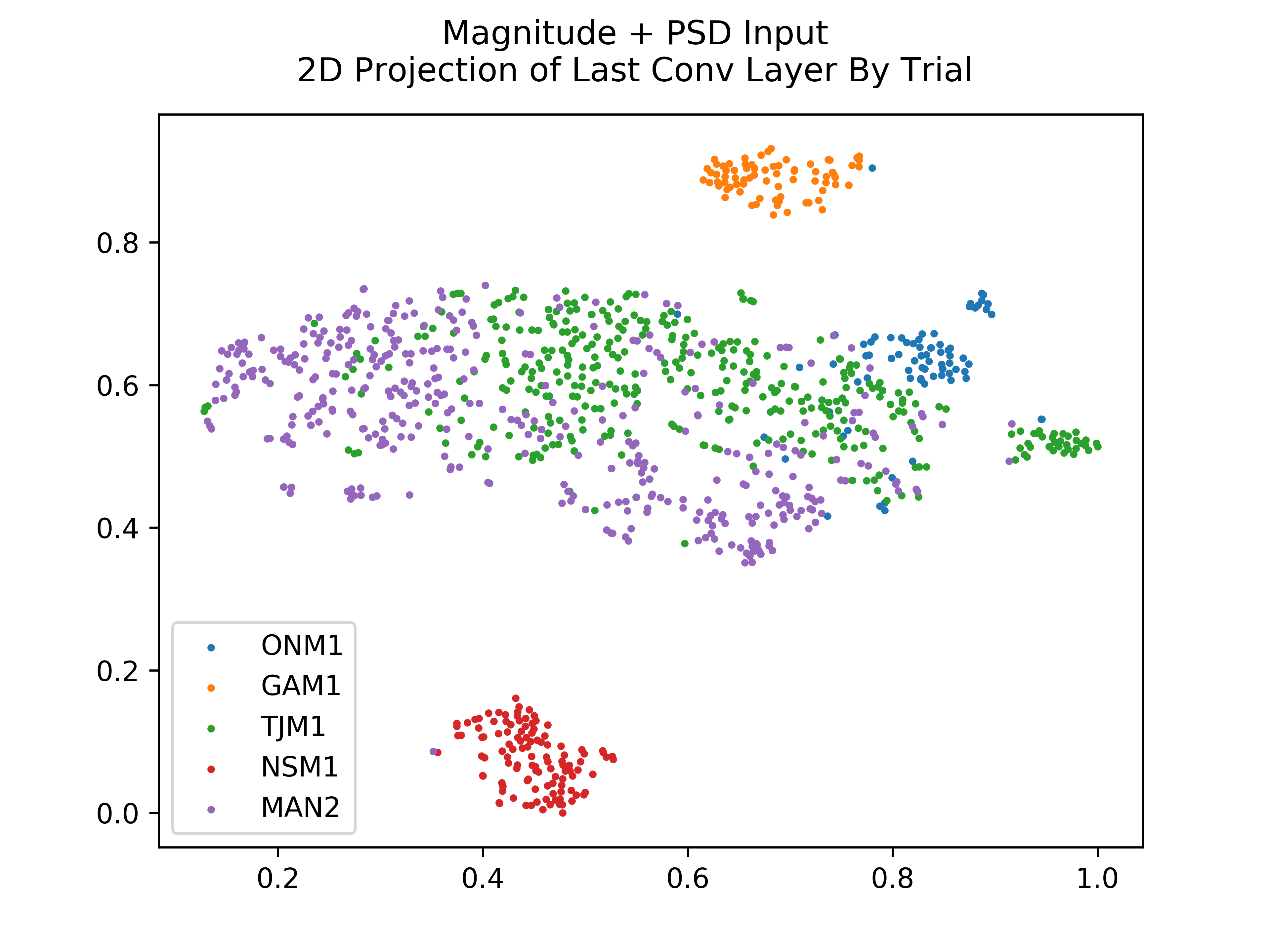}	
	\caption{Input sample points arranged in t-SNE projected feature space, color-coded by trial. Top is Magnitude-only network, middle is PSD-only network, and bottom is Magnitude+PSD network.}
	\label{fig:tsne_trial}
\end{figure}

We wondered how the feature space organization would change with the inclusion of additional representations. Figure \ref{fig:tsne_mag_mag_psd} shows the feature arrangement of the Magnitude + PSD net. We see a similar organization as previous but with clusters formed. We searched for possible meanings of the clustering.  Ultimately, we found the clustering correlated well to the trial name. This was unexpected for two reasons: (1) the training/test splits were not split over trial, and (2) the trial name was never used during the training procedure. We went back to the Magnitude-only input to determine if this organization was present and it was not.  The clustering phenomena was most prevalent in the Magnitude + PSD net. Figure \ref{fig:tsne_trial} depicts this clustering phenomena for three network configurations.

Currently, we are uncertain of the meaning of this unsupervised clustering though we have several hypotheses:
\begin{enumerate}
	\item The clustering reflects different sediment types which happen to correspond to different trials.
	\item The clustering is dependent on bottom texture.
	\item The clustering is a product of signal processing choices during the beamforming process which are trial dependent.  For example, signal gain or matched-filter differences.
\end{enumerate}

\section{Conclusion}
We created a simple convolutional neural network (CNN) classifier and measured the performance of it using various inputs from a synthetic aperture sonar (SAS) single-look complex (SLC) image.  Usually the phase information of the SLC is discarded as part of the classification procedure -- we evaluated its use in this work. Additionally, we applied a pre-trained off-the-shelf (OTS) network trained on photographs to SAS magnitude images using transfer learning. We demonstrated two ways to enhance SAS classifier performance when using CNNs: (1) utilization of the 2D power spectral density (PSD) derived from the normally discarded phase information, and (2) using a pre-trained OTS network trained on photographs.  We demonstrated these improvements using statistical testing to mitigate performance differences due to the stochastic nature of CNN training.  Finally, we analyzed the network internals to improve our understanding of the learning procedure.  We learned that the first layer convolutional weights are human interpretable for all input representations and that the mutual information between feature spaces is increased when different representations are trained in tandem versus exclusively. 

\clearpage 

\section{Acknowledgments}
ID Gerg would like to thank CMRE for hosting him during the time of this work and Joonho Park of PSU-ARL for the helpful discussions on 2D phase unwrapping and information theory. This work was partially supported by the Strategic Environmental Research and Development Program (SERDP) and by the NATO Allied Command Transformation (ACT).

\printbibliography

\end{document}